\documentclass[conference]{IEEEtran}
\usepackage{cite}
\usepackage{amsmath,amssymb,amsfonts}
\usepackage{algorithmic}
\usepackage{graphicx}
\usepackage{textcomp}
\usepackage{xcolor}
\usepackage{multirow}
\usepackage{bigstrut}
\usepackage{subfigure}
\usepackage[numbers,sort&compress]{natbib}
\def\BibTeX{{\rm B\kern-.05em{\sc i\kern-.025em b}\kern-.08em
    T\kern-.1667em\lower.7ex\hbox{E}\kern-.125emX}}
\usepackage{fancyhdr}

\begin{document}

\title{\huge AMMASurv: Asymmetrical Multi-Modal Attention for Accurate Survival Analysis with Whole Slide Images and Gene Expression Data\\
}

\author{\IEEEauthorblockN{Ruoqi Wang,
Ziwang Huang, Haitao Wang and
Hejun Wu}
\IEEEauthorblockA{School of Computer Science and Engineering,
Sun Yat-sen University\\
\{wangrq29, huangzw26, wanght39\}@mail2.sysu.edu.cn, wuhejun@mail.sysu.edu.cn}}

\maketitle

\thispagestyle{fancy}
\fancyhead{}
\lhead{}
\lfoot{}
\cfoot{}
\rfoot{}

\begin{abstract}
The use of multi-modal data such as the combination of whole slide images (WSIs) and gene expression data for survival analysis can lead to more accurate survival predictions. Previous multi-modal survival models are not able to efficiently excavate the intrinsic information within each modality. Moreover, previous methods regard the information from different modalities as similarly important so they cannot flexibly utilize the potential connection between the modalities. To address the above problems, we propose a new asymmetrical multi-modal method, termed as AMMASurv. Different from previous works, AMMASurv can effectively utilize the intrinsic information within every modality and flexibly adapts to the modalities of different importance. Encouraging experimental results demonstrate the superiority of our method over other state-of-the-art methods.
\end{abstract}


\section{Introduction}
The use of multi-modal data such as the combination of whole slide images (WSIs) and gene expression data for survival analysis can lead to more accurate survival predictions. As far as we know, there are few previous multi-modal methods that integrate WSIs and structured data to predict survival e.g. DeepCorrSurv \cite{yao2017deep} and MultiSurv \cite{vale2021long}. Structured data refer to data that can reside in a fixed field within a record or file like spreadsheets and gene expression data are a kind of structured data. When using WSIs and gene expression, there are two limitations in these approaches. 
Firstly, they are not able to efficiently detect the intrinsic information within each modality. For example, these methods just randomly select some patches from WSIs and use the patch features, ignoring making use of the integral information in WSIs. Moreover, they directly encode high-dimensional gene expression data that contain a lot of irrelevant information, introducing a lot of noise \cite{ioannidis2005microarrays, van2009survival}.
Secondly, they cannot flexibly utilize the potential connection between the modalities because they regard the information from different modalities as of equal importance even when the importance of modalities is different.

In this paper, we propose a method called Asymmetrical Multi-Modal Attention for Survival Analysis (AMMASurv) to solve the challenges. Our model is an end-to-end model that can incorporate information across modalities of different importance for the subsequent prediction task. It is built upon Transformer \cite{attention2017}, consisting of a asymmetrical multi-modal Transformer encoder followed by an multi-layer perception (MLP) \cite{dosovitskiy2021image}. We design an asymmetrical multi-modal attention (AMMA) mechanism in the encoder to fuse information from different modalities unevenly. With the AMMA which is unbalanced for different modalities, the information from noisy and less important modalities are not directly encoded. Instead, we provide directed inter-modality information transfer (only from important modality to unimportant modality) in AMMA, using the information from the more important modality to induce the representation of less important modality. In addition, AMMA provides intra-modality information transfer just in WSI modality to learn the correlation and interaction among WSI patches so that the integral information of WSIs can be learned. On the contrary, the intra-modality information transfer in gene expression data is prevented to preclude amplifying the influence of noise.

The main contributions of our work are summarized as follows:
\begin{enumerate}[]
\item An effective model for multi-modal survival prediction is proposed. Our method outperforms other state-of-the-art methods on two widely-used datasets. Furthermore, we study the influence of the individual modules in our method by ablation experiment.
\item Our model can solve the challenges in detecting the intrinsic information within each modality. It can learn the integral information of WSIs and denoise the gene expression representation. 
\item Our model can flexibly utilize the potential connection between the modalities by considering that the information from the two modalities is of different importance for survival prediction.

\end{enumerate}

\begin{figure*}[t]
\centerline{\includegraphics[height=5cm]{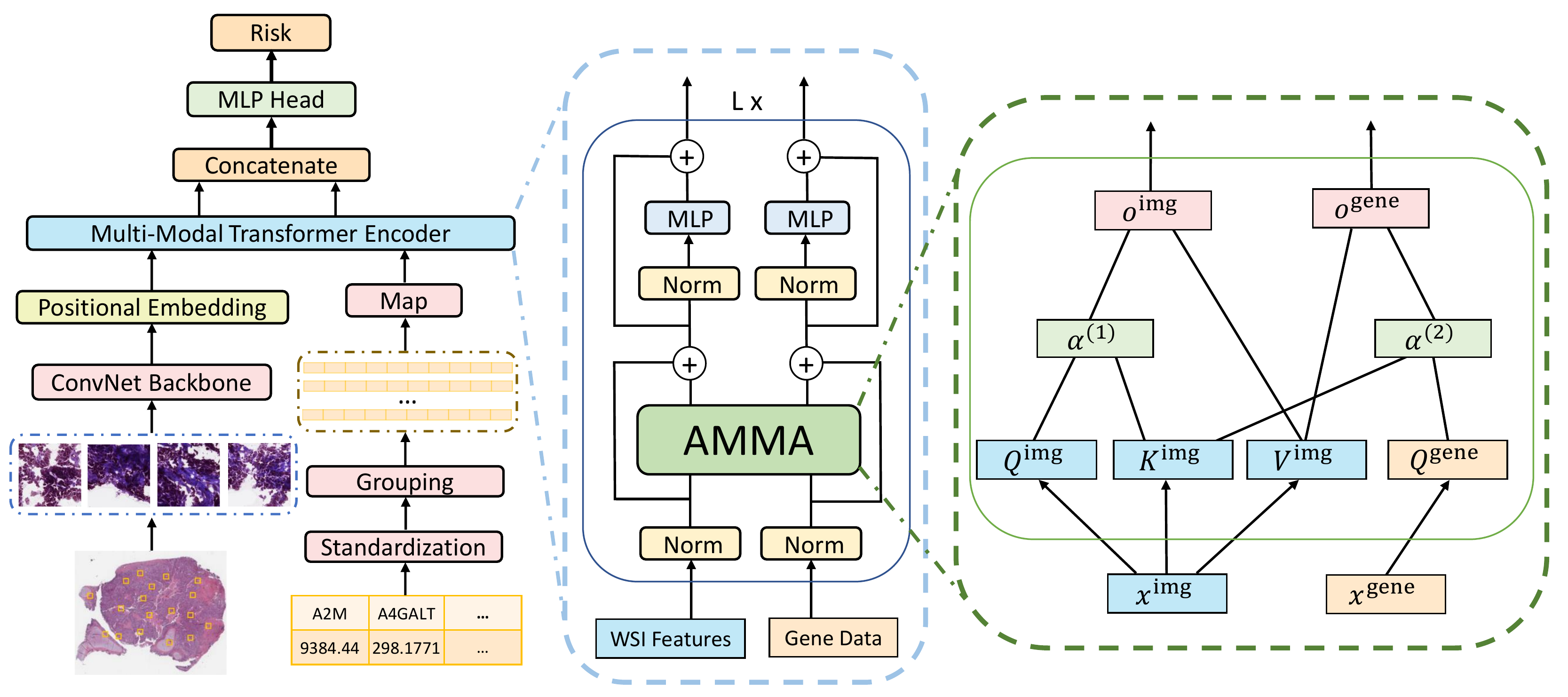}}
\caption{An overview of the proposed model. The left part is the overall process including feature extraction, feature incorporating and survival prediction. The middle part indicates the detailed structure of the multi-modal Transformer. The right part is the schematic diagram of AMMA.}
\label{modelfig}
\end{figure*}

\section{Methods}

In this paper, we propose an end-to-end model called Asymmetrical Multi-Modal Attention for Survival analysis (AMMASurv) to incorporate information across modalities of different importance for the subsequent prediction task. Our model is built upon Transformer \cite{attention2017}, consisting of a multi-modal encoder followed by an MLP \cite{dosovitskiy2021image}. The focus of our method is to build an Asymmetrical Multi-Modal Attention (AMMA) mechanism in the encoder to fuse information from two modalities efficiently. We show the overview of our model in Fig.~\ref{modelfig}.

\subsection{Image Features}

Inspired by SeTranSurv \cite{Huang2021integration}, we randomly select patches from each WSI and use a convolutional neural network backbone to extract the features of the patches. For a WSI as input, we randomly select $n$ patches of size $H \times W \times C$ from the foreground area, forming a sequence of patches $I\in \mathbb{R}^{n\times(H \times W \times C)}$. Then the convolutional neural network backbone $\operatorname{B}$ is applied on $I$ to extract a sequence of features $F_p\in \mathbb{R}^{n\times d_1}$ where 
\begin{align}
F_p=\operatorname{B}(I)\label{eqB}
\end{align}
and $d_1$ is the dimension of extracted features of a patch. In our model, the structure of pre-trained ResNet18 \cite{he2016deep} is utilized as the backbone network.

Same as SeTranSurv \cite{Huang2021integration}, we append position embeddings to the features $F_p$ to retain positional information of each patch. The process for generation positional vectors is the same as SeTranSurv \cite{Huang2021integration} and the dimension of a position vector is $d_{pos}$. Corresponding positional sequence $P\in {\mathbb{R}^{n\times d_{pos}}}$ is concatenated with $F_p$ to form $Z_0\in {\mathbb{R}^{n\times d}}$ where $d=d_1+d_{pos}$. Finally, similar to class token in BERT \cite{devlin2019bert}, we also perpend a learnable embedding $Z_{token}^0$, whose state at the output of the Transformer encoder $Z_{token}^L$ can serve as the WSI representation $y_1$. Finally, we concatenate $z_{token}^0$ and $Z_0$ into $Z_1$\eqref{eqconcat}.
\begin{align}
Z_1=[Z_{token}^0;Z_0]\label{eqconcat}
\end{align}

\subsection{Gene Expression Features}

Firstly, we standardize the data of $N$ gene symbols and segment the data into $m$ groups with $d_2 = N/m$ gene symbols in each group, forming a gene feature sequence $G\in {\mathbb{R}^{m\times d_2}}$ where $d_2<d_1$ for each patient according to the dataset. To facilitate subsequent calculations, we expand $G$ to $F_g\in \mathbb{R}^{m \times d_1}$ by repeating the values in $G$ and then concatenate $F_g$ and zero vectors which are of the same dimension as positional embedding vector in $P$ to get $F_g'\in \mathbb{R}^{m\times d}$. Finally we utilize a layer of learnable fully connected neural network $M$ and non-linear function $\operatorname{ReLU}$ \cite{nair2010rectified} to map $F_g'$ into $Z_2\in {\mathbb{R}^{m\times d}}$ where 
\begin{align}
Z_2 = \operatorname{ReLU}(M(F_g')). \label{z2}
\end{align}

\subsection{AMMA}
AMMA is the core of our model. In order to explain it more clearly, we firstly introduce the principle of attention mechanism from the perspective of graph and then explain the details of AMMA. 

\subsubsection{Principle of AMMA}

Inspired by \cite{yao2020multimodal}, the method of incorporating features from different modalities is based on a graph perspective of Transformer. Incorporating information from another modality is similar to adding other different nodes and edges into the original graph. Therefore, to joint information from both WSIs and gene data, we build a graph that contains heterogeneous nodes including nodes of WSI features and nodes of gene features. Different from traditional self-attention in Transformer, our graph is not fully connected because the two modalities are not equally important. Instead, there are not any outgoing arcs from the gene expression nodes which are less important, meaning that the noisy representation of gene data cannot influence other features at all but can update themselves by the induction from WSI features.

\subsubsection{Details of AMMA}

In order to achieve the effect described above, we design AMMA. In AMMA, each hidden representation of gene data is induced from features of WSI patches and can update itself under the guidance of attention. A visual representation is shown on the right of Fig.~\ref{modelfig}.

Formally, we consider the input representation as $x^\mathrm{img}\in \mathbb{R}^{n\times d}$ and $x^\mathrm{gene}\in \mathbb{R}^{m\times d}$, the output of multi-modal self-attention is computed as follows:
\begin{align}
o^{\mathrm{img}}_{i}=\sum_{j=1}^{n} \alpha^{(1)}_{i j}\left(x_{j}^{\mathrm{img}} W^{V}\right)\label{oimgi}
\end{align}
\begin{align}
o^{\mathrm{gene}}_{i}=\sum_{j=1}^{n} \alpha^{(2)}_{i j}\left(x_{j}^{\mathrm{img}} W^{V}\right)\label{ogenei}
\end{align}
where $\alpha ^{(1)}_{ij}$ and $\alpha ^{(2)}_{ij}$ is the weight coefficient computed by a softmax function, respectively:
\begin{align}
\alpha^{(1)}_{i j}=\operatorname{softmax}\left(\frac{\left(x^\mathrm{img}_{i} W^{Q}\right)\left(x_{j}^{\mathrm{img}} W^{K}\right)^{T}}{\sqrt{d}}\right)\label{alpha1}
\end{align}
\begin{align}
\alpha^{(2)}_{i j}=\operatorname{softmax}\left(\frac{\left(x^\mathrm{gene}_{i} W^{Q}\right)\left(x_{j}^{\mathrm{img}} W^{K}\right)^{T}}{\sqrt{d}}\right)\label{alpha2}.
\end{align}
And $o^\mathrm{img}\in \mathbb{R}^{n\times d}$ \eqref{oimgi} and $o^\mathrm{gene}\in \mathbb{R}^{m\times d}$ \eqref{ogenei} are the hidden representation of WSI patches and gene expression data. In AMMA, the updating of hidden representation of the gene expression features is induced from WSIs under the guide of attention.

\subsection{Multi-Modal Transformer Encoder}
As shown in the middle of Fig.~\ref{modelfig}, the Multi-Modal Transformer Encoder consists of multiheaded asymmetrical multi-modal attentions (AMMA) and MLP blocks. The AMMA can calculate the relation among features of different patches in WSI and the relation between each WSI patch and each gene expression group. Layernorm (LN) is applied before every block and residual connection \cite{he2016deep} is utilized after every block. The MLP contains two layers with a GELU \cite{hendrycks2016gaussian} non-linearity. Using the superscript $l$ to indicate the state of the feature representation of each modality in layer $l=1,...,L$, the encoding process is as follows:

\begin{align}
Z_1^{'l}=\operatorname{AMMA}(\operatorname{LN}(Z_1^{l-1}),\operatorname{LN}(Z_2^{l-1}))+Z_1^{l-1}\label{z1pl}
\end{align}
\begin{align}
Z_2^{'l}=\operatorname{AMMA}(\operatorname{LN}(Z_1^{l-1}),\operatorname{LN}(Z_2^{l-1}))+Z_2^{l-1}\label{z2pl}
\end{align}
\begin{align}
Z_1^{l}=\operatorname{MLP}(\operatorname{LN}(Z_1^{'l}))+Z_1^{'l} \label{ztl}
\end{align}
\begin{align}
Z_2^{l}=\operatorname{MLP}(\operatorname{LN}(Z_2^{'l}))+Z_2^{'l}\label{zt2}
\end{align}
\begin{align}
y_1=\operatorname{LN}(Z_{token}^L) \label{y1}
\end{align}
\begin{align}
y_2=\operatorname{LN}(\operatorname{MeanPooling}(Z_2^L)) \label{y2}
\end{align}
\begin{align}
y=[y_1;y_2] \label{y}
\end{align}

\subsection{Survival Prediction}

The output of multi-modal Transformer encoder $y_1$ and $y_2$ are concatenated into $y$ which represents the joint features of two madalities. It goes through an MLP Head module \cite{dosovitskiy2021image} and directly generates predicting risk $R$ \eqref{R} where $W^{(1)}$ and $W^{(2)}$ refer to learned weight.  
\begin{align}
R=W^{(2)}\operatorname{ReLU}(W^{(1)}y)\label{R}
\end{align}

If a patient has multiple WSIs, we compute the joint features of each WSI and gene data of this patient and average the risk scores of all the predictions to get the final risk score. 

The loss function is negative Cox log partial likelihood \cite{zhu2017wsisa} for censored survival data.

\section{Experiments}
To verify the proposed model, we conduct a series of experiments including comparison with other state-of-the-art methods and ablation study. The details are shown in the following subsections.

\subsection{Dataset Description}
We use two datasets from The Cancer Genome Atlas (TCGA) \cite{kandoth2013mutational} which provides high-resolution WSIs and gene expression data to verify the validity of our model.  We conduct experiments on two cancer types respectively: Lung squamous cell carcinoma (LUSC) and Ovarian serous cystadenocarcinoma (OV). We select patients who have both WSIs and gene expression data and use them in experiments. In the experiments, we perform five-fold cross-validation on each dataset.

Following \cite{Huang2021integration}, the WSIs and gene expression data are used without RoI annotations, so we compare it with the state-of-the-art methods using the same datasets without RoI annotations.

\subsection{Comparison with State of the Art Methods}
\subsubsection{Baselines}
We compare our model with eight models that can be divided into three categories i.e. mono-modal survival models using only WSIs, mono-modal survival models using only structured data and multi-modal survival models using both WSIs and structured data.

\begin{table}[t]
  \centering
  \caption{Performance comparison of our model and other methods using C-index values on LUSC and OV datasets.}
    \begin{tabular}{c|c|c|c}
    \hline
    Data  & Model & LUSC  & OV \\
    \hline
    \hline
    \multirow{2}[6]{*}{WSI-only} & RankSurv\cite{di2020ranking}       & 0.674    & 0.667 \\
\cline{2-4}                  & DeepAttnMISL\cite{yao2020whole}   & 0.670     & 0.659 \\
\cline{2-4}                  & SeTranSurv\cite{Huang2021integration}     & \textbf{0.701} & \textbf{0.692} \\
    \hline
    \hline
    \multirow{2}[6]{*}{Gene expression-only} & FCN-Surv\cite{yao2017deep} & 0.559 & 0.556 \\
\cline{2-4}                             & MSR-RF\cite{wright2017unbiased} & 0.567 & \textbf{0.561} \\
\cline{2-4}                             & Survival SVM\cite{polsterl2015fast} & \textbf{0.607} & 0.543 \\
    \hline
    \hline
    \multirow{2}[6]{*}{WSIs \& Gene expression} & DeepCorrSurv\cite{yao2017deep} & 0.563      &  0.619 \\
\cline{2-4}                         & MultiSurv\cite{mariotto2014cancer} & 0.605 & 0.609 \\
\cline{2-4}                         & \textbf{AMMASurv} & \textbf{0.759} & \textbf{0.745} \\
    \hline
    \end{tabular}
  \label{table2}
\end{table}

\subsubsection{Results and discussion}
To assess the performance of the models in survival prediction, we use the concordance index (C-index) as the evaluation metric \cite{steck2008ranking}.
The results of our model and the baselines are shown in Table~\ref{table2}. Our model outperforms any other previous works for both ovarian and lung cancer by incorporate information from two modalities.

\subsection{Ablation Study}
\begin{figure}[t]
\centerline{\includegraphics[height=4.5cm]{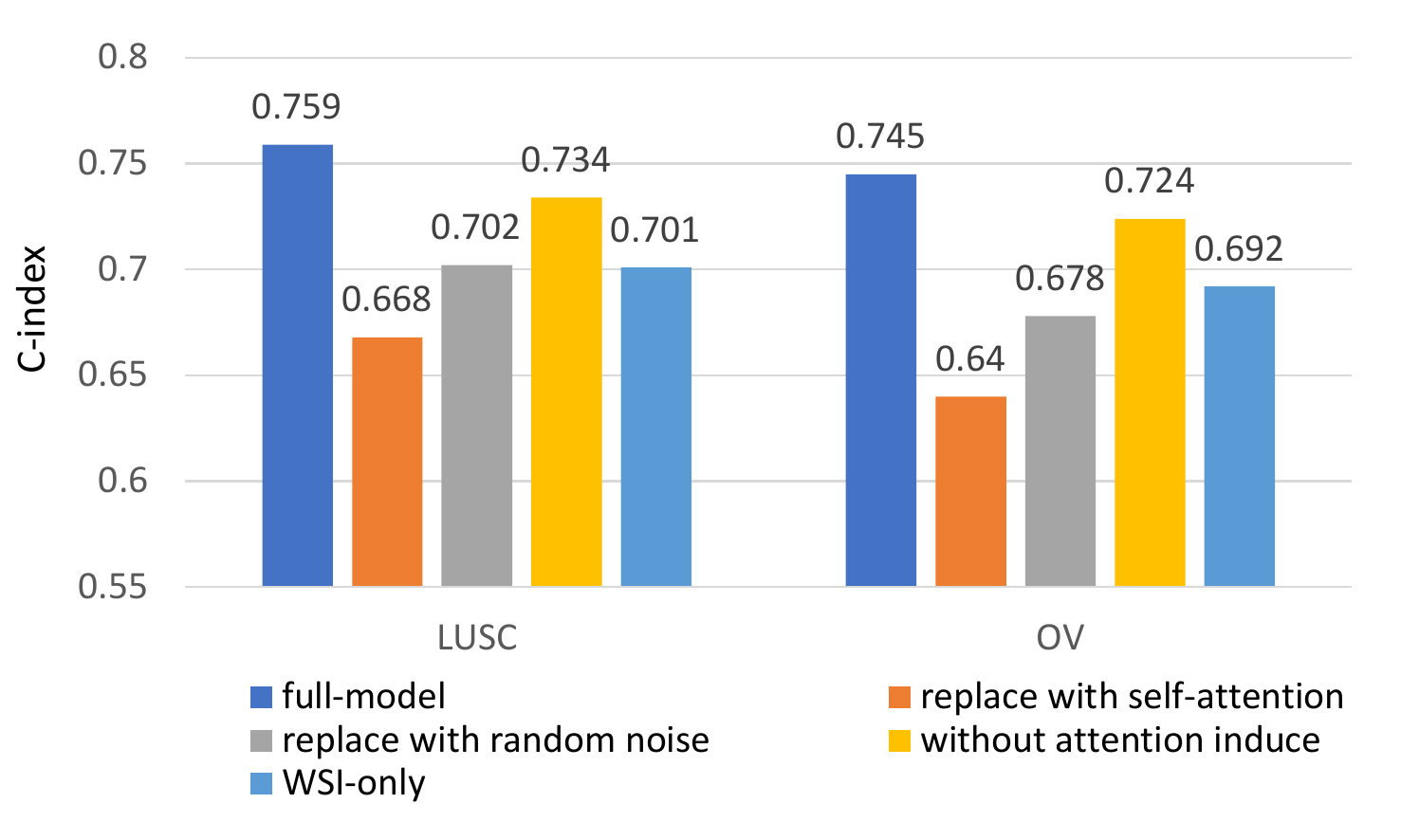}}
\caption{Ablation results of different components in our model. The left one is the results on the LUSC dataset and the right one is the results on the OV dataset.}
\label{ablares}
\end{figure}

To further study the influence of the individual modules in our method, we conduct ablation experiments and the results are presented in Fig.~\ref{ablares}.

Firstly, inspired by \cite{yao2020multimodal}, we replace the AMMA with traditional self-attention. The changed model performs even worse than the WSI-only model, indicating that our model can avoid the interference of noise in gene expression data successfully.

Secondly, inspired by \cite{yao2020multimodal,elliott2018adversarial}, we replace all input gene expression features with random vectors. The model performs similarly to or even worse than the WSI-only models, showing that the helpful information for survival in gene expression can be excavated and used by our model.

Finally, we fix the representation of gene expression and concatenate the uninduced gene representation directly to the encoded WSI representation directly. The experimental results show the importance of the inducing mechanism.

\section{Conclusion}
In this paper, we propose AMMASurv to incorporate information from modalities of different importance for survival prediction using WSIs and gene expression data. AMMASurv efficiently excavates the intrinsic information within each modality and flexibly utilizes the potential connection between the modalities. Experimental results show the superiority of our method over other state-of-the-art methods.

\bibliographystyle{IEEEtran}
\bibliography{IEEEabrv,mylib}

\end{document}